# Short Text Conversation Based on Deep Neural Network and Analysis on Evaluation Measures

Hsiang-En Cherng, Chia-Hui Chang
Nation Central University
Department of Computer Science and Information Engineering
Taoyuan, Taiwan
seancherng.tw@gmail.com, chia@csie.ncu.edu.tw

*Abstract*—With the development of Natural Language Processing, Automatic question-answering system such as Waston, Siri, Alexa, has become one of the most important NLP applications. Nowadays, enterprises try to build automatic custom service chatbots to save human resources and provide a 24-hour customer service. Evaluation of chatbots currently relied greatly on human annotation which cost a plenty of time. Thus, [32] has initiated a new Short Text Conversation subtask called Dialogue Quality (DQ) and Nugget Detection (ND) which aim to automatically evaluate dialogues generated by chatbots. In this paper, we solve the DQ and ND subtasks by deep neural network. We proposed two models for both DQ and ND subtasks which is constructed by hierarchical structure: embedding layer, utterance layer, context layer and memory layer, to hierarchical learn dialogue representation from word level, sentence level, context level to long range context level. Furthermore, we apply gating and attention mechanism at utterance layer and context layer to improve the performance. We also tried BERT to replace embedding layer and utterance layer as sentence representation. The result shows that BERT produced a better utterance representation than multi-stack CNN for both DQ and ND subtasks and outperform other models proposed by other researches. The evaluation measures are proposed by [26], that is, NMD, RSNOD for DQ and JSD, RNSS for ND, which is not traditional evaluation measures such as accuracy, precision, recall and f1-score. Thus, we have done a series of experiments by using traditional evaluation measures and analyze the performance and error.

I. INTRODUCTION

Automatic question-answering system has become one of the most important applications in Natural Language Processing. For example, Waston, a question-answering system developed by IBM, competed on quiz show *Jeopardy!* and won the first place prize of $1 million in 2011. Siri, a virtual assistant developed by Apple Inc, uses voice queries and a natural-language user interface to answer questions, make recommendations, and perform actions by delegating requests. Alexa, a virtual assistant developed by Amazon which is capable of voice interaction, making to-do lists, provide real-time information and control smart devices. Nowadays, enterprises try to build automatic customer service chatbots to replace human in customer service. With customer service chatbots, customer service departments are able to save a plenty of time and human resources and provide a 24-hour chatbot to answer customers' questions.

Effective evaluation measures are necessary to evaluate the quality of chatbots, due to the few methods to automatically evaluate such systems, it spends a plenty of time to evaluate by human. In this paper, we discuss the Short Text Conversation task from [32] to show several methods that evaluate the quality and structure of dialogue between a chatbot and a customer. With such measures, the quality of chatbots could be evaluated automatically and efficiently without human annotation.

The goal of DQ is to evaluate the quality of the whole dialogue in three measures: Task Accomplishment (A-score), to evaluate whether the problem has been solved, and solve to what extent. Dialogue Effectiveness (E-score), to judge whether the utterers interact effectively to solve the problem efficiently. Customer Satisfaction of the dialogue (S-score), to evaluate the customer satisfaction of this dialogue, not the product or the company. Instead of traditional evaluation measures, the evaluation of DQ subtask is based on Normalized Match Distance (NMD) and Root Symmetric Normalized Order-Aware Divergence (RSNOD) as defined in [26].

ND can be considered as a kind of Dialog Act (DA) labeling problem. Most researches consider DA labeling problem as sequence labeling problem and use traditional machine learning methods [1][16][20]. Recently, many deep learning models [6][10][19][28][30][31] are proposed to tackle the problem. The golden answer of DA labeling datasets: Switchboard Dialog Act Corpus (SWDA) [4] and The ICSI Meeting Recorder Dialog Act (MRDA) [5] are a certain tag. However, the golden answer of ND is the utterance's nugget probability distribution instead of a certain nugget tag, thus the evaluation is based on Jensen-Shannon Divergence (JSD) and Root Normalized Sum of Squared (RNSS) scores which measure the probability distribution between outputs and golden answers as defined in [26]. With such golden answer, we could not directly apply traditional measures for sequence labeling such as accuracy, precision, recall and f1-score.

In this paper, we proposed several DNN models based on hierarchical structure with embedding layer, utterance layer, context layer, memory layer and output layer. With hierarchical structure, dialogue representation could be learned in different level from word, sentence to context, which is similar with real world dialogues that helps the model to learn dialogue representation more effectively. The idea of such hierarchical structure is from [6], but we apply deeper model by adding multi-stack techniques and memory enhance structure to capture rich n-gram information and long range context features between utterances. Moreover, we apply gating mechanism to control whether to drop or to keep the features generated by CNN which helps the model to generalize easier. For DQ subtask, we used word2vec with gated multi-stack CNN as sentence representation and 1-stack gated CNN as context representation. After that, we apply memory network [8] to further capture context information between utterances by using self-attention mechanism. For ND subtask, since the lack of training data, we remove gating mechanism and memory layer to avoid overfitting. In context layer, we replace 1-stack gated CNN with multi-stack BI-LSTM which results in better performance in DQ subtask.

In this paper, we first discuss the related work of dialogue After that, in section 3 and 4, we introduce our proposed models and the evaluation measures for both short text conversation DQ and ND subtask, compare the performance between our models, participants' models and baseline models and analyze the learning curve of different size of training data to see if our models could result in better performance with more training data. Finally, in section 5, we give a short conclusion of this paper.

## II. RELATED WORK

In this section, we discuss the related work of precious researches on dialogue, including question-answering system, dialogue system and short text conversation

### A. Question Answering System

Question-answering system aims to select an answer for a given question, there are several question-answering datasets such as Microsoft Machine Reading Comprehension Dataset (MS MACRO) [18], Stanford Question Answering Dataset 1.1 (SQuAD 1.1) [21], SQuAD 2.0 [22]. Most of the models solving question-answering tasks using deep learning methods. In this section, we discuss the previous researches using deep learning methods on these three datasets.

[12] proposed a model called Masque model (Multi-style Abstractive Summarization model for Question answering) contains question reader, passage ranker, answer classifier and answer sentence decoder. This complex model is built by transformer blocks, attention mechanism and take Glove and ELMo as text representation, achieved state-of-the-art performance on MS MACRO 2.1. [17] designed a deep attention-based multi-task model for machine reading comprehension which evolves from document and paragraph level candidate text to more precise answer extraction. [25] proposed a special model called novel Memory Augmented Machine Comprehension Network (MAMCN) model by Bi-GRU encoders and a memory controller to utilize external memory to compensate for the limited memory capacity of the recurrent layers to better reason over long documents. [15] introduced one model and two techniques for machine reading comprehension tasks, the reinforced mnemonic reader model, re-attention mechanism and dynamic-critical reinforcement learning. With these proposed mechanisms, the model could refine current attentions to predict a more acceptable answer.

### B. Dialogue System

Dialogue generation aims to build a dialogue system by generating corresponding responses for given input questions. There are three types of dialogue system using different ideas on dialogue generation: Rule-based (template-based) model, retrieval-based model and generative model.

Previous researches on rule-based and template-based dialogue generation systems have been around for a long time [23][14], which highly rely on human designed rules to generate responses. These systems are able to give precise responses if the input question meet the pre-defined rules. However, these systems do not scale with increasing domain complexity and maintaining the increasing number of templates, which could not deal with complex questions such as logical reasoning or troubleshooting.

Retrieval-based dialogue system requires a knowledge base with large amount of question-answer pairs, then respond to the input question by the most similar question-answer pair in the knowledge base. Retrieval-based model without machine learning [3][1] could be considered as standard baselines which sometimes perform well on dialogue tasks.

Nowadays, Researchers focus on machine learning methods to build dialogue systems, all rules and components could be trained from the given dialogues themselves, escape the limitation. [7] proposed an adversarial training method for open-domain dialogue generation which generator takes a form similar to seq2seq models and discriminator is a binary classifier. [9] proposed a new text generation model, Diversity-Promoting Generative Adversarial Network (DP-GAN), to avoid producing repeated and boring expressions in existing text generation methods. DP-GAN encourage the generator to generate fluent text instead of repeated text which can generate more diverse and informative text than baselines. [13] proposed an special model: Auto-Encoder Matching (AEM) model which attention mechanism to learn the utterance-level semantic dependency and the relation between input and output. The AEM model generates responses of high coherence and fluency compared to baseline models in DailyDialog dataset [29]. [27] proposed a conditional variation auto-encoders which learns potential conversational intents and generates responses using greedy decoders.

### C. Short Text Conversation

Both DQ and ND subtasks in this research could be consider as short text conversation problems. Previous researches on short text conversation have investigated different techniques including: Hidden Markov Model [2], Naïve Bayes [20], Conditional Random Fields [2][20][16], and deep learning [6][10][19][28][30][31]. Early deep learning models rely on CNN and BI-LSTM modules [10]. The CNN-based model outperforms the BILSTM-based model in both Switchboard Dialog Act Corpus (SWDA) dataset [4] and Meeting Recorder Dialog Act (MRDA) [5] dataset. Hierarchical CNN and BI-LSTM models are latter proposed to better represent sentences [19]. For example, [30] applied hierarchical CNN and hierarchical BI-LSTM for sentence and dialogue representation for DA labeling. More recently, CRF-based DNN model such as LSTM+CRF models are proposed in [31][28]. Furthermore, combining hierarchical BI-LSTM structure with CRF layer to represent utterance and dialogue is also studied in [6]. The major difference of the ND subtask to traditional DA labeling is the output: for each utterance, the ground truth is not a single label but label distribution in DQ subtask. Thus, the performance evaluation is based on JSD and RNSS instead of traditional measures such as accuracy, precision, recall and f1-score.

### III. DIALOGUE QUALITY SUBTASK

The goal of DQ subtask is to evaluate the quality of a dialogue by three measures: Task Accomplishment (A-score), Dialogue Effectiveness (E-score) and Customer Satisfaction of the dialogue (S-score). We proposed two models for DQ subtask, the major difference of these two models is sentence representation, embedding layer and utterance layer, one is based on skip-gram with multi-stack CNN and the other is based on BERT structure.

*A. Model*

In this section, we followed the idea of hierarchical CNN in [30] to construct our model, memory enhance hierarchical multi-stack CNN with gating mechanism (MeHGCNN), but use multi-stack CNN. We apply 2-stack CNN to utterance representation and 1-stack CNN to context representation. There are 5 layers in our proposed model including input embedding layer, utterance layer, dialog context layer, memory layer and output layer (as Figure 1). The goal of such hierarchical structure is to decode the input dialogue hierarchically from word, sentence to context, to capture the dependency of words and utterances.

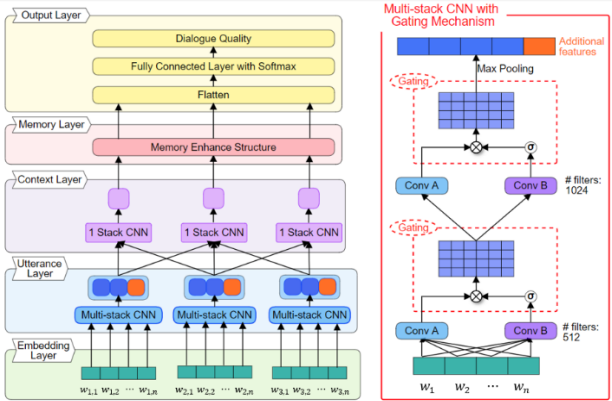

Figure 1. Memory enhance hierarchical gated CNN (MeHGCNN)

First, for embedding layer, we apply a word2vec skip-gram model with dimension 100 trained by wiki text8 and DQ&ND dialogue dataset [32].

Second, after embedding layer, we send the tokens in utterance to utterance layer, we Use 2-stack gated CNN as utterance representation. Gating mechanism is implemented by element-wise multiplication and sigmoid function between the output of two convolution operations. Using gating mechanism in multi-stack CNN could generalize features effectively. Let $X_i = [w_{(i,1)}, w_{(i,2)}, \ldots, w_{(i,n)}]$ where $X_i$ denotes the $ith$ utterance and $w_{(i,1)}, w_{(i,2)}, \ldots, w_{(i,n)}$ denotes the token in $ith$ utterance. 2-stack gated CNN is defined as follows:

$$ulA_i = ConvA(X_i) \quad (1)$$
$$ulB_i = ConvB(X_i) \quad (2)$$
$$ulC_i = ulA_i \odot \sigma(ulB_i) \quad (3)$$
$$ul_i = [maxpool(ulC_i), speaker_i, nugget_i] \quad (4)$$

$ulA_i$ denotes the features generated by convolution A, $ulB_i$ denotes the masks (or gates) generated by convolution B and $ulC_i$ denotes the utterance representation after applying gating mechanism between $ulA_i$ and $ulB_i$ where $\odot$ denotes element-wise multiplication and $\sigma$ denotes sigmoid function. For 2-stack CNN, we execute (1) to (3) for twice. Finally, we apply max-pooling operation and add additional speaker and nugget features for the output of Utterance Layer as (4).

Third, in context layer, we apply 1-stack gated CNN. The structure of gated CNN is same as utterance layer but only 1-stack. We take the concatenation of [previous utterance, current utterance, next utterance] as input to learn the dependency between adjacent utterance.

Fourth, since context layer could only capture the dependency between adjacent utterances, we follow the idea of memory network to further capture long range context information. The overview of memory layer is shown as Figure 2. We need to prepare Input Memory and Output Memory by Bi-GRU as (5) to (10).

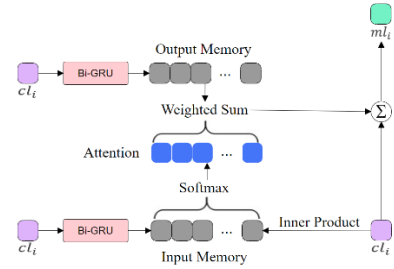

Figure 2. Overview of memory layer

$$\overrightarrow{I_i} = \overrightarrow{GRU}(cl_i, h_{i-1}) \quad (5)$$
$$\overleftarrow{I_i} = \overleftarrow{GRU}(cl_i, h_{i+1}) \quad (6)$$
$$I_i = tanh(\overrightarrow{I_i} + \overleftarrow{I_i}) \quad (7)$$
$$\overrightarrow{O_i} = \overrightarrow{GRU}(cl_i, h_{i-1}) \quad (8)$$
$$\overleftarrow{O_i} = \overleftarrow{GRU}(cl_i, h_{i+1}) \quad (9)$$
$$O_i = tanh(\overrightarrow{O_i} + \overleftarrow{O_i}) \quad (10)$$

Where $cl_i$ is the output of context layer for the $ith$ utterance and $h_{i-1}, h_{i+1}$ are the hidden states for Bi-GRU. $I_i$ and $O_i$ denote the Input Memory and Output Memory respectively. Next, we calculated the attention weight $w_i$ by the inner product between Input Memory $cl_i$ as (11) and calculated the weighted sum between Output Memory and the attention weight as (12). Finally, we concatenate the output for all utterance as (13) as the output of memory layer.

$$w_i = \frac{exp(cl_i \cdot I_i)}{\sum_{i'=1}^{k} exp(cl_{i'} \cdot I_{i'})} \quad (11)$$

$$ml_i = \sum_{i'=1}^{k} w_{i'} \cdot O_{i'} + cl_i \quad (12)$$

$$ml = [ml_1, ml_2, ..., ml_k] \quad (13)$$

The final layer of our model is output layer, we take the concatenated output from memory layer as input, then apply simple fully connected layer with softmax as (14) and (15) to output the distribution for A-score, E-score and S-score.

$$fc = mlW_{fc} + b_{fc} \quad (14)$$

$$P(score|dialogue) = \frac{exp(fc_i)}{\sum_{i'=1}^{5} exp(fc_{i'})} \quad (15)$$

### B. Experiments

*1) Dataset:* The DQ and ND subtask datasets are dialogues as the corpus [32]. The training data contains 1,672 dialogues with a total of 8,672 utterances, validation data are randomly selected from training data. Testing data contains 390 dialogues with a total of 1,755 utterances. The label of both DQ and ND subtasks is annotated by 19 students from the department of Computer Science, Waseda University.

*2) Evaluation measures:* The goal of DQ subtask is to predict the dialogue quality for a given dialogue in A-score, E-score and S-score. Since the result of DQ subtask is non-nominal distribution, we need to apply cross-bin measures to utilize the distance between bins instead of bin-by-bin measures. In evaluation, we compare the predicted dialogue quality distribution with the golden answer by two cross-bin measures, *Normalized Match Distance (NMD)* and *Root Symmetric Normalized Order-Aware Divergence (RSNOD)* as defined in [26].

*3) Comparison of our proposed models:* This section shows the performance in NMD and RSNOD. *Table 1* shows the performance of Simple BERT, a model with only BERT embedding and output layer; MeHGCNN, the model we discuss on section 3.A; and MeGCBERT, replace the embedding layer and utterance layer with BERT sentence embedding. The result shows that BERT embedding improves the performance comparing with word2vec. But even BERT is a powerful language model, it still need context information to gain better result. Moreover, shows the ablation of MeGCBERT, the result shows that three techniques we used, gating mechanism, memory layer and adding nugget information as additional features, are useful in improving the performance.

Table 1. Performance of our proposed model for DQ subtask

| Model | (A-score) | | (E-score) | | (S-score) | |
|---|---|---|---|---|---|---|
| | N | R | N | R | N | R |
| Simple BERT | 0.093 | 0.138 | 0.089 | 0.134 | 0.084 | 0.134 |
| MeHGCNN | 0.086 | 0.131 | 0.081 | 0.122 | 0.079 | 0.124 |
| **MeGCBERT** | **0.082** | **0.126** | **0.079** | **0.120** | **0.076** | **0.125** |

Table 2. Ablation of MeGCBERT

| Model | (A-score) | | (E-score) | | (S-score) | |
|---|---|---|---|---|---|---|
| | N | R | N | R | N | R |
| **MeGCBERT** | **0.082** | **0.126** | **0.079** | **0.120** | **0.076** | **0.125** |
| w/o gating | 0.089 | 0.132 | 0.081 | 0.121 | 0.082 | 0.129 |
| w/o memory | 0.091 | 0.014 | 0.081 | 0.124 | 0.080 | 0.127 |
| w/o nuggets | 0.096 | 0.139 | 0.080 | 0.120 | 0.078 | 0.125 |

*4) Comparison with other works:* This section shows the model performance comparing with other works. BL-popularity is a baseline model which predicts the probabililty of the other labels except the most popular label as 0, BL-uniform is a baseline model which always predict the uniform distribution and BL-lstm is a baseline model leverages Bi-LSTM, all of these models were proposed by [32]. CUIS is a model based mo hierarchical attention network with BERT embedding [11] and SLSTC-run0 to run2 are models based on BI-LSTM with multi-head attention, multi-task learning and BERT embedding, respectively [24]. The result shows that our model outperforms all other's proposed models.

Table 3. Model comparison of DQ subtask

| Model | (A-score) | | (E-score) | | (S-score) | |
|---|---|---|---|---|---|---|
| | N | R | N | R | N | R |
| BL-popularity | 0.185 | 0.253 | 0.195 | 0.278 | 0.150 | 0.233 |
| BL-uniform | 0.168 | 0.248 | 0.158 | 0.216 | 0.199 | 0.268 |
| BL-lstm | 0.090 | 0.132 | 0.082 | 0.122 | 0.084 | 0.131 |
| CUIS | 0.090 | 0.136 | 0.085 | 0.128 | 0.084 | 0.135 |
| SLSTC-run0 | 0.102 | 0.149 | 0.094 | 0.140 | 0.091 | 0.142 |
| SLSTC-run1 | 0.091 | 0.139 | 0.086 | 0.132 | 0.082 | 0.134 |
| SLSTC-run2 | 0.093 | 0.137 | 0.083 | 0.122 | 0.082 | 0.131 |
| **MEGCBERT** | **0.082** | **0.126** | **0.079** | **0.120** | **0.076** | **0.125** |

### C. DQ as a triditional classification problem

Since the DQ subtasks use label probability distribution as training and testing data, we are not able to evaluate model performance by intuitive measures such as accuracy. In this section, we convert the label distribution of testing data to one-hot labeling and consider the DQ subtask as a traditional classification problem. In evaluation part if there are two or more labels then evaluate our models in accuracy. Table 6 shows the performance of MeHGCNN and MeGCBERT. The result of MeHGCNN is better than MeGCBERT which is conflict with the result of NMD and RSNOD. The reason of such conflict is that although most of the result generated by MeGCBERT is similar with the golden answer distribution, the score with the highest probability is not always same as golden answer. On the other hand, MeHGCNN is good at prediction the highest score label, but the score distribution between golden answer is not as similar as the one created by MeGCBERT, which result in the high accuracy but low NMD, RSNOD. Table 4 shows an example of dialogue, both MEHGCNN and MEGCBERT recognize the correct highest A-Score "1", but when we consider the distribution, MeGCBERT is much better than MEHGCNN which result in the better NMD and RSNOD. Table 5 shows another example, even MeGCBERT fail to predict the score with highest probability, the distribution is still similar with the golden answer that result in low NMD and RSNOD.

Table 4. An example of DQ dialogue

| Utterances | A-Score NMD distribution | | | | | |
|---|---|---|---|---|---|---|
| **Customer:** I want to use the data in SIM card 2. How do I set up? What's wrong with it? @Smartisan Technology Customer Service @Luo Yonghao<br>**Helpdesk:** Hi, currently you can't choose to use the data from SIM card 1 or SIM card 2 with a shortcut key….<br>**Customer:** The dialogue box popped up while I chose SIM card 2. I can't set up.<br>**Helpdesk:** Hi, please try to re-install the Micro SIM card in the card slot 2.<br>**Customer:** Still can't.<br>**Helpdesk:** If convenient, you can check if there is any after-sale points nearby in Offline Maintainance Store - Smartisan Technology….<br>**Customer:** OK. Thank you! | Source | -2 | -1 | 0 | 1 | 2 |
| | Golden Answer | 0.053 | 0.158 | 0.263 | 0.474 | 0.053 |
| | MeHGCNN | 0.026 | 0.038 | 0.119 | 0.442 | 0.374 |
| | MeGCBERT | 0.106 | 0.080 | 0.270 | 0.455 | 0.089 |

Table 5. Another example of DQ dialogue

| Utterances | A-Score NMD distribution | | | | | |
|---|---|---|---|---|---|---|
| **Customer:** When the cellphone USB with type c interface is inserted into M1L, why is there no display... Does it need the operation of other USB debugging?...<br>**Helpdesk:** H Hello! M1L cellphone doesn't support OTG function. The charging interface can't recognize external equipment.<br>**Customer:** Are all Smartisan cellphones not suitable for OTG external memory function ... How about the cellphones released newly?<br>**Helpdesk:** H Smartisan Pro and Smartisan Pro 2 both support OTG function. If you are urgent to use the data in the USB, you are suggested to export the data to the computer and then copy it to the cellphone.<br>**Customer:** OK, Thank you!. | Source | -2 | -1 | 0 | 1 | 2 |
| | Golden Answer | 0.000 | 0.105 | 0.053 | 0.368 | 0.474 |
| | MeHGCNN | 0.026 | 0.022 | 0.075 | 0.413 | 0.464 |
| | MeGCBERT | 0.044 | 0.015 | 0.086 | 0.430 | 0.425 |

Table 6. Performance of DQ as a traditional classification problem

| Model | (A-score) Accuracy | (E-score) | (S-score) |
|---|---|---|---|
| **MeHGCNN** | **0.710** | **0.662** | **0.777** |
| MeGCBERT | 0.659 | 0.659 | 0.728 |

## IV. NUGGET DETECTION SUBTASK

### A. Model

The goal of ND subtask is to classify the nugget type for all utterances in a given customer-service dialogue. There are seven types of nugget for each utterance as shown in Table 7. We proposed two hierarchical models for ND subtask, the major difference is sentence representation, one is based on skip-gram + multi-stack CNN and the other is based on BERT structure.

Table 7. Nuggets for ND subtask

| Nugget | Description |
|---|---|
| CNUG0 | **Customer trigger:** Problem stated |
| CNUG* | **Customer goal:** Solution confirmed |
| CNUG | **Customer regular:** Utterances contain information to solution |
| CNaN | **Customer Not-a-nugget:** Utterances do not contain information to solution |
| HNUG* | **Helpdesk goal:** Solution stated |
| HNUG | **Helpdesk regular:** Utterances contain information to solution |
| HNaN | **Helpdesk Not-a-nugget:** Utterances do not contain information to solution |

In this section, we followed the idea of hierarchical CNN in [30] to construct our model, multi-stack CNN with LSTM, but use 3-stack CNN to represent an utterance. For context representation, we apply 2-stack Bi-directional LSTM (as Figure 3). There are 4 layers in our proposed model including input embedding layer, utterance layer, dialog context layer, and output layer. The embedding layer is same as the one for DQ subtask.

The utterance layer, instead of gated CNN, we apply two convolution operations with different filter size to capture different n-gram features, concatenate both of the output, and apply max-pooling operation that is:

$$ulA_i = ConvA(X_i) \quad (16)$$

$$ulB_i = ConvB(X_i) \quad (17)$$

$$ulC_i = [ulA_i, ulB_i] \quad (18)$$

$$ul_i = [maxpool(ulC_i), speaker_i] \quad (19)$$

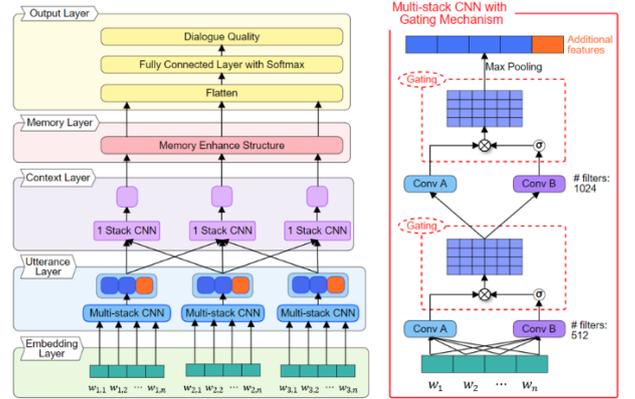

Figure 3. Hierarchical CNN + BI-LSTM (HCNN-LSTM)

For context layer, we simply use 2-stack Bi-LSTM layer to capture features between adjacent utterances. Since there is no memory layer in HCNN-LSTM, we need 2-stack structure to capture rich n-gram features in sentence level.

$$\overrightarrow{cl_i} = \overrightarrow{LSTM}(ul_i, h_{i-1}) \quad (20)$$

$$\overleftarrow{cl_i} = \overleftarrow{LSTM}(ul_i, h_{i+1}) \quad (21)$$

$$cl_i = tanh\left(\overrightarrow{cl_i^l} + \overleftarrow{cl_i^l}\right) \quad (22)$$

Where $h_{i-1}$ and $h_{i+1}$ denote the hidden states of Bi-LSTM, we apply tanh activation function to combine the features generated by forward and backward LSTM. Finally, after 2-stack operation, the output of context layer for the $ith$ utterance is denoted as $cl_i$.

Finally, for output layer, we output the nugget probability distribution for all utterance by softmax activation function. Since the evaluation measures of ND is the loss between the golden answer and the prediction. We could only apply softmax activation function instead of CRF as output layer. The function of output layer is as follows:

$$(nugget|u_i) = \frac{exp(Wcl_i)}{\sum_{i'=1}^{k} exp(cl_{i'})} \quad (23)$$

*B. Experiments*

*1) Dataset:* The dataset for ND subtask is same as the one for DQ subtask.

*2) Evaluation measures:* The goal of ND subtask is to predict the nugget distribution for all utterances in a given dialogue. For evaluation, we compare the predicted nugget distribution with the golden answer by two bin-by-bin measures, *Jensen-Shannon Divergence (JSD)* and *Root Normalized Sum of Squares (RNSS)* as defined in [26], which simply accumulate the error in each bin between two normalized distribution.

*3) Comparison of our proposed models*: This section shows the performance in NMD and RSNOD of our proposed models as shown in Table 8 .BERT-LSTM is our proposed model HCNN-LSTM but replace the embedding layer and utterance layer as BERT sentence embedding. BERT-LSTM outperform HCNN-LSTM which denotes that BERT is more powerful than word2vec + multi-stack CNN as sentence representation. With the comparison between BERT-LSTM and simple BERT, it shows that context layer is also important for ND subtask since BERT-LSTM outperform simple BERT model. Moreover, *Table 9* shows the ablation of BERT-LSTM for ND subtask. The result shows that 2-stack BI-LSTM outperforms 1-stack BI-LSTM. As Table 10, adding gating mechanism, or memory layer doesn't help the model to improve the performance. It might be result from the insufficient training data that cause underfitting in complex models. As the learning curve analysis in Figure 4, Both JSD and RNSS reduce when adding number of training data until 100% training data are used. This tendency shows our model could perform better if there is more training data for ND subtask. On the other hand, we cannot expect a complex model to perform well without sufficient training data, this is the major reason that we do not apply gating mechanism and memory layer in our proposed models for ND subtask.

Table 8. Performance of our proposed model for ND subtask

| Model | JSD | RNSS |
|---|---|---|
| Simple BERT | 0.0341 | 0.1171 |
| HCNN-LSTM | 0.0246 | 0.0962 |
| **BERT-LSTM** | **0.0228** | **0.0933** |

Table 9. Ablation of BERT-LSTM

| Model | JSD | RNSS |
|---|---|---|
| **BERT-LSTM** | **0.0228** | **0.0933** |
| W/o multi-stack | 0.0246 | 0.0951 |

Table 10. Gating and memory experiments

| Model | JSD | RNSS |
|---|---|---|
| **BERT-LSTM** | **0.0228** | **0.0933** |
| W/ gating | 0.0244 | 0.0960 |
| W/ memory layer | 0.0234 | 0.0941 |

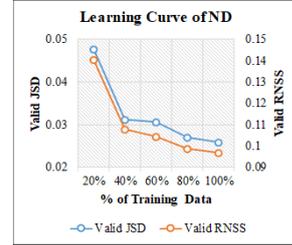

Figure 4. Learning Curve of ND

*4) Comparison with other works:* The model comparison with other researches, which is same as the DQ subtask, is shown as . Our BERT-LSTM model results in the best performance in JSD and RNSS comparing with all other models.

Table 11. Gating and memory experiments

| Model | JSD | RNSS |
|---|---|---|
| BL-popularity | 0.1665 | 0.2653 |
| BL-uniform | 0.2304 | 0.3709 |
| BL-lstm | 0.0248 | 0.0952 |
| SLSTC-run0 | 0.0289 | 0.1037 |
| SLSTC-run1 | 0.0252 | 0.0973 |
| SLSTC-run2 | 0.0263 | 0.0979 |
| **MeGCBERT** | **0.0228** | **0.0933** |

*C. ND as a traditional sequence labeling problem*

Since the ND subtasks use label probability distribution as training and testing data, we could only apply softmax layer instead of CRF layer to predict label distribution. Also, we are not able to evaluate model performance by intuitive measures such as precision, recall and f1-score. In this section, we convert the label distribution of both training data and testing data to one-hot labeling, consider the ND subtask as a traditional sequence labeling problem then solve the ND subtask by CRF instead of softmax and show the performance in precision, recall and f1-score.

*1) Preprocessing:* Converting probability distribution label to one-hot encoding label, we consider the nugget with highest probability as the true label. However, there might be two nuggets with the highest probability in one utterance. In that case, we create two one-hot encoding labels for the dialogue. For instance, a nugget distribution for all utterances in a dialogue as Table 12, utterance 4 has the two highest probability HNUG* and HNUG, so we create two one-hot labels for this dialogue: [CNUG0, HNUG, CNUG, HNUG*] and [CNUG0, HNUG, CNUG, HNUG]. As another example, if there are two utterances with two highest nugget probability (as Table 13), we create for one-hot encoding labels, that is: [CNaN, HNUG, CNUG, HNaN], [CNaN, HNaN, CNUG, HNaN], [CNaN, HNUG, CNaN, HNaN], [CNaN, HNaN, CNaN, HNaN]. In testing phase, if an utterance has two highest nugget probabilities, we consider both of them are the correct answer. Table 15 shows the number of training data, and testing data after preprocessing.

Table 12. One dialog with one utterance having two highest nugget probability

| Utt No. | CNUG0 | CNUG* | CNUG | CNaN | HNUG* | HNUG | HNaN |
|---|---|---|---|---|---|---|---|
| 1 | 0.947 | 0 | 0 | 0.053 | 0 | 0 | 0 |
| 2 | 0 | 0 | 0 | 0 | 0.053 | 0.894 | 0.053 |
| 3 | 0.053 | 0 | 0.894 | 0.053 | 0 | 0 | 0 |
| 4 | 0 | 0 | 0 | 0 | 0.474 | 0.474 | 0.052 |

Table 13. One dialog with two utterances having two highest nugget probability

| Utt No. | CNUG0 | CNUG* | CNUG | CNaN | HNUG* | HNUG | HNaN |
|---|---|---|---|---|---|---|---|
| 1 | 0.421 | 0 | 0.105 | 0.474 | 0 | 0 | 0 |
| 2 | 0 | 0 | 0 | 0 | 0.052 | 0.474 | 0.474 |
| 3 | 0.052 | 0 | 0.474 | 0.474 | 0 | 0 | 0 |
| 4 | 0 | 0 | 0 | 0 | 0 | 0.421 | 0.579 |

Table 14. Confusion matrix of nuggets

| Nugget | CNUG* | CNUG | CNaN | CNUG0 | HNUG* | HNUG | HNaN |
|---|---|---|---|---|---|---|---|
| CNUG* | 19 | 16 | 10 | 0 | 0 | 0 | 0 |
| CNUG | 9 | 431 | 43 | 1 | 0 | 0 | 0 |
| CNaN | 3 | 23 | 57 | 0 | 0 | 0 | 0 |
| CNUG0 | 0 | 0 | 12 | 374 | 0 | 0 | 0 |
| HNUG* | 0 | 0 | 0 | 0 | 27 | 14 | 2 |
| HNUG | 0 | 0 | 0 | 0 | 17 | 619 | 31 |
| HNaN | 0 | 0 | 0 | 0 | 0 | 21 | 70 |

Table 15. # data after preprocessing

| # Dialogues (# Utts) | Original | After preprocessing |
|---|---|---|
| Training data | 1,337 (5,601) | 1,467 (6,150) |
| Testing data | 335 (1,316) | 368 (1,461) |

*2) Result: Table 16* shows the performance of ND subtask as a traditional sequence labeling problem, both HCNN-LSTM and BERT-LSTM models perform in accuracy higher than 88.8%. Except macro recall, BERT-LSTM model outperforms HCNN-LSTM model in other three measurements. However, even the overall accuracy is nearly 90%, the macro average of precision, recall and f1-measure are only 75% to 85%, this result shows that there might be several nuggets that are hard to be classified. *Table 17* shows the performance of each nugget type, the f1-score of CNUG*, CNaN, HNUG* and HNaN are lower than 72.4%. The result shows that it's hard for models to recognize the goal of customer and helpdesk (CNUG* / HNUG*) and Non-a-nugget (CNaN / HNaN). Furthermore, we analyze the confusion matrix of BERT-LSTM model as *Table 14* where rows represent the predictions and columns represent the true labels. We found that there are several nugget pairs that are hard to classify: [CNUG, CNaN], [CNUG*, CNUG], [CNUG*, CNaN], [HNUG*, HNUG] and [HNUG, HNaN]. Since the true labels is from 19 annotators, we also analyze the average probability difference between two highest nuggets in an utterance of human annotations as Table 18. The lower difference between two nuggets means the higher chance that annotators hold different opinion, vice versa. As Table 18, the average probability difference of [CNUG, CNaN], [CNUG*, CNUG], [CNUG*, CNaN], [HNUG*, HNUG] and [HNUG, HNaN] are lower than 0.46, which means it's also difficult for human annotators to recognize these nuggets correctly.

Table 16. Performance of ND as a traditional sequence labeling

| Model | Accuracy | Macro P | Macro R | Macro F |
|---|---|---|---|---|
| HCNN-LSTM | 88.8% | 75.6% | **74.8%** | 75.2% |
| BERT-LSTM | **89.9%** | **83.4%** | 74.6% | **78.7%** |

Table 17. P/R/F for each nugget

| Nugget | Precision | Recall | F-score |
|---|---|---|---|
| CNUG* | 76.9% | 32.3% | 45.5% |
| CNUG | 87.2% | 95.3% | 91.1% |
| CNaN | 75.8% | 56.6% | 64.8% |
| CNUG0 | 98.4% | 99.7% | 99.1% |
| HNUG* | 100.0% | 25.0% | 40.0% |
| HNUG | 89.8% | 98.8% | 94.1% |
| HNaN | 88.7% | 61.2% | 72.4% |

Table 18. Avg prob difference between 2 highest nuggets

| Nugget pair | Avg prob diff | # pairs | percentage |
|---|---|---|---|
| CNUG0, CNUG* | 0.842 | 13 | 0% |
| CNUG0, CNUG | 0.731 | 242 | 3% |
| CNUG0, CNaN | 0.696 | 1,508 | 22% |
| CNUG*, CNUG | 0.348 | 232 | 3% |
| CNUG*, CNaN | 0.339 | 36 | 1% |
| CNUG, CNaN | 0.455 | 1,793 | 26% |
| HNUG*, HNUG | 0.307 | 865 | 13% |
| HNUG*, HNaN | 0.118 | 8 | 0% |
| HNUG, HNaN | 0.401 | 2,220 | 32% |

## V. CONCLUSION

In this paper, we proposed two hierarchical multi-stack models for both DQ and ND subtasks. The experiments show that multi-stack mechanism, gating mechanism and memory enhance structure could improve all three types of score in our proposed model MeHGCNN for DQ subtask. However, applying gating mechanism and memory enhance structure drop the performance in ND subtask since the lack of training data which might cause underfitting. Moreover, we also try BERT as sentence representation which outperforms the performance of all measures in both DQ and ND subtasks comparing with other works on the same task. Finally, since DQ and ND subtask use special evaluation measures, we also worked on the analysis of applying traditional evaluation methods such as accuracy, precision, recall and f1-score.